\documentclass{svproc}
\usepackage[utf8]{inputenc}
\usepackage{url}
\usepackage{helvet}         % selects Helvetica as sans-serif font
\usepackage{courier}        % selects Courier as typewriter font
\usepackage{amssymb}        % for \mathbb and other math symbols
\usepackage{color}
\usepackage{makeidx}         % allows index generation
\usepackage{graphicx}        % standard LaTeX graphics tool
\usepackage[misc]{ifsym}                            % when including figure files
\usepackage{multicol}        % used for the two-column index
\usepackage[bottom]{footmisc}% places footnotes at page bottom
\usepackage{tabularx}
\usepackage{multirow}
\usepackage{natbib}
\usepackage{algorithm}
\usepackage{algorithmic}
\usepackage[center]{caption}
\usepackage{longtable}
\usepackage{rotating}
 \usepackage{amsmath}
\usepackage{subfigure}

\begin{document}
\mainmatter              % start of a contribution
\title{Explainability in Generative Medical Diffusion Models: A Faithfulness-Based Analysis on MRI Synthesis}
\titlerunning{Explainability in Generative Medical Diffusion
Models}  % abbreviated title (for running head)
\author{Surjo Dey \and Pallabi Saikia \thanks{Corresponding author}}
\authorrunning{Surjo and Pallabi} % abbreviated author list (for running head)
% \tocauthor{Ivar Ekeland, Roger Temam, Jeffrey Dean, David Grove,
% Craig Chambers, Kim B. Bruce, and Elisa Bertino}
%
\institute{Rajiv Gandhi institute of petroleum technology, Amethi, Uttar Pradesh, India\\
\email{22cs3072@rgipt.ac.in}, \email{psaikia@rgipt.ac.in} }

\maketitle              % typeset the title of the contribution

\begin{abstract}
% This study investigates the explainability of generative diffusion models in the context of medical imaging, focusing on Magnetic resonance imaging (MRI) synthesis. Although diffusion models have shown strong performance in generating realistic medical images, their internal decision making process remains largely opaque. We present a faithfulness-based explainability framework that analyzes how prototype-based explainability methods like ProtoPNet (PPNet), Enhanced ProtoPNet (EPPNet), and ProtoPool can link the relationship between generated and training features. Our study focuses on understanding the reasoning behind image formation through denoising trajectory of diffusion model and subsequently prototype explainability with faithfulness analysis. Experimental analysis shows that EPPNet achieves the highest faithfulness (with score 0.1534), offering more reliable insights, and explainability into the generative process. The results highlight that diffusion models can be made more transparent and trustworthy through faithfulness-based explanations, contributing to safer and more interpretable applications of generative AI in healthcare. 

Diffusion models have demonstrated strong capability in generating realistic medical images, however their internal decision making process still remains largely unclear. In this study, we investigated the explainability of generative diffusion models in the context of medical imaging, with a specific focus on Magnetic resonance imaging (MRI) synthesis. We present a faithfulness-based explainability framework to examine how prototype-based explainability methods like ProtoPNet (PPNet), Enhanced ProtoPNet (EPPNet), and ProtoPool can capture the relationship between generated and training features. Furthermore, the study investigates the reasoning behind the image formation through denoising trajectory of diffusion model and subsequently prototype explainability with faithfulness measure. Experimental analysis across different prototype-based approaches show that EPPNet able to achieve the highest faithfulness (with score 0.1534), offering more reliable insights, and explainability into the generative process. The results highlight that diffusion models can be made more transparent and trustworthy through faithfulness-based explanations, contributing to safer and more interpretable applications of generative AI in healthcare.

\keywords{Explainability, Medical Image synthesis, Diffusion model, MRI, Prototype Learning}
\end{abstract}

\section{Introduction}\label{sec:intro}
Diffusion based generative models have recently become one of the most popular ways to create high-quality images. These models learn by a gradually noise adding process, so they can turn random noise into a clear and realistic image step by step \cite{DDPM} \cite{sde}. Diffusion models, when compared with other generative approaches like GANs \cite{GAN} or VAEs \cite{VAE}, are more stable and useful in areas such as medical imaging \cite{kazerouni2023diffusion}. There are many limitations of these models to use in healthcare, such as, the lack of strong convergence guarantees when working with complex and noisy data, and limited explainability. Moreover, the models are often unclear how the generated images relate to the training data, and what influences specific visual features during the generation process. This lack of transparency reduces trust and make it difficult to apply in clinical settings.

Recent studies have begun exploring the integration of explainability mechanisms into diffusion-based generative models for medical imaging. Recent works, including VALD-MD (2024) \cite{siddiqui2024vald} and Latent Diffusion Attribution (2024) \cite{siddiqui2025latent} employs latent diffusion to reconstruct normal counterparts of abnormal scans, providing voxel-level attributions for diagnostic insight. While The Mediffusion framework (2024) \cite{kaleta2024mediffusion} jointly trains diffusion and classification objectives to achieve self-explainable semi-supervised generation. Unlike prior multi-class attribution studies, we propose a prototype-based faithfulness framework for single-class medical diffusion that directly associates generated features with their real training origins. These advances demonstrate a growing trend toward faithful and interpretable diffusion frameworks in healthcare, motivating the present study’s prototype-based faithfulness analysis for explainable MRI synthesis. Although explainability has been widely explored in classification or diagnostic models, it has received far less attention in generative systems. In particular, understanding how diffusion models create new medical images, such as MRI scans still remains an open question \cite{Explainibility}.

This work investigates the prototype-based explainability for generative diffusion models that synthesize breast MRI images for segmentation. Our key contributions include:

\begin{itemize}
    \item We proposed a faithfulness-based explainability framework that integrates prototype-based reasoning with diffusion-based medical image generation model.
    \item We analyzed three prototype-based explainability models (PPNet, EPPNet, ProtoPool) to systematically quantify the interpretability of diffusion-generated MRI images.
    \item The models are evaluated on different Image qualiy analysis metrics PSNR, SSIM, LPIPS to analyse the visual quality of diffusion synthesised images and Faithfulness Score metric is applied for analysing the alignment of prototype activations with generative dynamics.
    \item The experiments are performed on the publicly available medical imaging dataset, the DUKE Breast MRI data set, demonstrating the reasoning for realistic synthesis of MRI data using diffusion model with explainable generation trajectories and faithfulness scores.
\end{itemize}

While many existing studies depends on multiple classes, our task involves only one single anatomical class, focusing purely on image level relationships. We evaluate how different prototype-based explainability methods can trace the influence of real training examples on generated images. Our analysis shows that prototype-based reasoning can provide faithful and interpretable explanations even in single class medical domains \cite{PPNet} \cite{diffonmri}. By connecting the image generation process with clear and meaningful explanations, we move toward making AI in medical imaging more transparent and reliable. 
%The implemented code is publicly available at: https://github.com/surjo0/Explainability

\section{Methodology}

This work looks at how to make diffusion-based generative models more explainable when they are used to generate MRI images. The methodology connect a diffusion model architecture for generating realistic MRI images with prototype based neural networks that provide interpretable mappings between generated and real data samples. The objective is to explore whether these models can offer faithful explanations of how specific image features emerge during the generative process. Our approach focuses on a single class medical imaging domain, where all data belong to the same anatomical category, ensuring that variations appear only from structural or textural differences rather than semantic class distinctions. By combining the stochastic generative dynamics of diffusion models with prototype learning, we aim to construct a framework that not only synthesizes anatomically accurate MRI images but also reveals which regions or training examples most influence the generated outputs.

\subsection{Diffusion Model Architecture for MRI Synthesis}

Diffusion models are generative frameworks that learn to reverse a gradual noising process, transforming random noise into structured images. In our context, the model is trained to generate realistic breast MRI images by progressively denoising a sample from a Gaussian distribution using Denoising Diffusion Probabilistic Models \cite{DDPM}. The learning objective is to approximate the reverse of a fixed forward diffusion process that slowly adds noise to real MRI data.

The forward diffusion process can be expressed as a Markov chain that adds Gaussian noise to an image $x_0$ over $T$ time steps. Each step produces a slightly noisier version $x_t$ according to:
\begin{equation}
q(x_t | x_{t-1}) = \mathcal{N}(x_t; \sqrt{1 - \beta_t}\, x_{t-1}, \beta_t \mathbf{I}),
\end{equation}
where $\beta_t$ represents the variance schedule controlling the noise level at step $t$. This process gradually destroys image information until $x_T$ becomes nearly pure noise.

To generate new samples, the model learns a reverse process parameterized by $\theta$:
\begin{equation}
p_\theta(x_{t-1} | x_t) = \mathcal{N}(x_{t-1}; \mu_\theta(x_t, t), \Sigma_\theta(x_t, t)),
\end{equation}
where $\mu_\theta$ and $\Sigma_\theta$ are predicted by a neural network, typically a U-Net or its variant. This network learns to estimate either the mean of the clean image or the noise added during the forward process.

Training aims to minimize the difference between the true noise $\epsilon$ added during diffusion and the model’s predicted noise $\epsilon_\theta(x_t, t)$:
\begin{equation}
\mathcal{L}_{\text{simple}} = \mathbb{E}_{x_0, \epsilon, t} \left[ \| \epsilon - \epsilon_\theta(x_t, t) \|^2 \right].
\end{equation}
This loss encourages the model to denoise accurately at each time step, effectively learning how to reconstruct realistic MRI images from noise. During inference, the model iteratively applies the learned reverse process, gradually converting random noise into a coherent breast MRI image.

To further improve anatomical consistency, the diffusion process is conditioned on additional information such as segmentation maps or structural priors \cite{int1}, \cite{int2}, \cite{int3}. In this setup, the model learns a conditional distribution $p_\theta(x_{t-1} | x_t, y)$, where $y$ encodes prior information about tissue boundaries or organ structure. This conditioning allows the generated MRI images to better preserve spatial alignment and realistic anatomical details.

\subsection{Prototype-Based Explainability Framework}

To make the generative diffusion model explainable, we employ a prototype-based framework that links generated MRI images to representative training samples. Prototype learning assumes that each model decision can be explained through a small set of learned image regions, called prototypes, which represent typical local patterns observed in the data \cite{PPNet}. When the model generates a new image, it can be interpreted as a combination of these prototypes, allowing us to trace how specific visual features emerge.

Prototype-based models differ from standard convolutional neural networks by including an additional layer that explicitly stores a set of prototype vectors $\{p_j\}_{j=1}^{m}$ in the latent feature space. These prototypes are compared with local feature patches extracted from an image through similarity measures such as the squared Euclidean distance. Given a feature map $f(x) \in \mathbb{R}^{H \times W \times D}$, the similarity between a feature patch $f(x)_{hw}$ and a prototype $p_j$ is computed as:
\begin{equation}
s_{j,hw} = - \| f(x)_{hw} - p_j \|_2^2,
\end{equation}
where higher similarity indicates that the local region resembles the prototype. The model then aggregates these similarities to form interpretable predictions or associations.

\begin{figure}[t]
\centering
\includegraphics[scale=0.60]{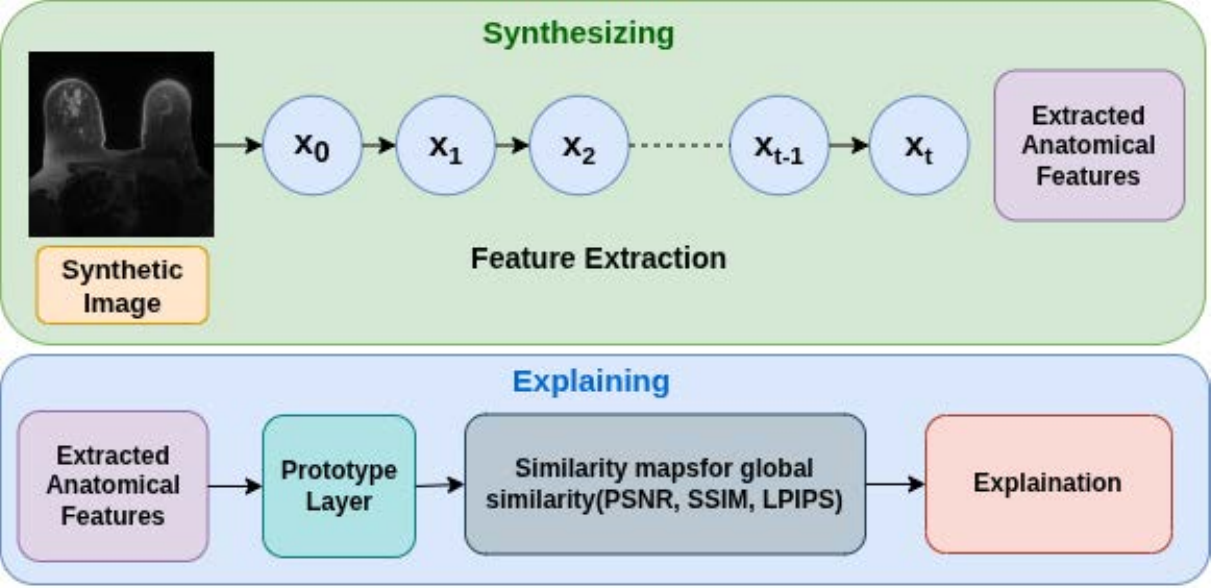}
\caption{Overall architecture of the proposed diffusion-based explainability framework. In the synthesizing stage, the model gradually extracting important anatomical features. In the explaining stage, these features are passed through a prototype layer to find similarities between real and generated MRI regions using metrics like PSNR, SSIM, and LPIPS. This helps provide a clear and interpretable explanation of how the model produces the images.}
\label{fig:method_architecture}
\end{figure}

In our study, we explore three main prototype-based architectures: ProtoPNet (PPNet) \cite{PPNet}, Enhanced ProtoPNet (EPPNet) \cite{EPPNet}, and ProtoPool \cite{ProtoPool}. Each provides a different mechanism for learning and updating prototypes, thereby offering distinct interpretability properties Figure \ref{fig:method_architecture}. 

\subsubsection{Prototype Propagation Network (PPNet)}

The PPNet forms the foundation of prototype based explainability. It introduces an interpretable layer that associates learned prototypes with specific regions in an image. Each prototype represents a feature patch extracted from the training set and acts as an example of a visual concept, such as a texture or structure within an MRI scan.

The model architecture typically follows a convolutional backbone (e.g., ResNet or U-Net encoder) followed by a prototype layer and a fully connected classification head. The output of the prototype layer is a similarity score between each prototype $p_j$ and the input feature map $f(x)$, computed as:
\begin{equation}
g_j(x) = \max_{h,w} s_{j,hw} = - \min_{h,w} \| f(x)_{hw} - p_j \|_2^2,
\end{equation}
where $s_{j,hw}$ represents the similarity between the prototype and the feature at position $(h,w)$. This allows each prototype to focus on the most relevant image region.

Training involves two objectives: classification accuracy and prototype alignment. The alignment loss encourages prototypes to remain close to the feature patches they represent:
\begin{equation}
\mathcal{L}_{\text{align}} = \sum_{j} \min_{x \in X_j} \| p_j - f(x)_{h^*, w^*} \|_2^2,
\end{equation}
where $X_j$ denotes the set of samples associated with prototype $p_j$. This measure ensures that prototypes correspond to real and meaningful regions from the training images with improved interpretability.

\subsubsection{Enhanced ProtoPNet (EPPNet)}

The EPPNet is based on the original PPNet wuth better prototype normalization and clearer interpretability scoring. This makes similarity scores more stable and helps provide reliable explanations of the generated data.

The key idea is to compute normalized influence scores that illustrate how strongly each prototype contributes to a given image. The normalized influence of prototype $p_j$ for image $x$ is defined as:
\begin{equation}
NIS_j(x) = \frac{\exp(g_j(x))}{\sum_{k=1}^{m} \exp(g_k(x))},
\end{equation}
where $g_j(x)$ is the similarity activation from the prototype layer. 

$NIS_j(x)$ converts similarity scores into probabilities and measure the contribution of each prototype to the overall representation.

EPPNet introduces a regularization term \text{div} that ensures prototypes remain diverse and capture different regions or structures. This diversity loss prevents prototypes from collapsing into similar patterns, thereby improving both visual coverage and interpretability. 

\begin{equation}
\mathcal{L}_{\text{div}} = \sum_{i \neq j} \exp(-\| p_i - p_j \|_2^2).
\end{equation}

Regularization helps EPPNet for more faithful explanations by highlighting which prototypes most influence a generated MRI image, linking synthetic features to their corresponding real patterns in the dataset.

\subsubsection{ProtoPool}

ProtoPool is an extension of prototype-based framework that introduces a dynamic pooling mechanism allowing the prototypes to be shared and merged across related samples. Unlike PPNet and EPPNet, which maintain fixed prototype assignments, ProtoPool organizes prototypes into a shared latent pool that can adaptively allocate prototypes based on feature similarity. The pooled representation in Protopool is computed as:
\begin{equation}
z_i = \sum_{j=1}^{m} \alpha_{ij} p_j, \quad \text{where} \quad \alpha_{ij} = \frac{\exp(-\| f(x)_i - p_j \|_2^2)}{\sum_{k=1}^{m} \exp(-\| f(x)_i - p_k \|_2^2)}.
\end{equation}

where, $\mathcal{P} = \{p_1, p_2, \dots, p_m\}$ denote the prototype pool, and $\alpha_{ij}$ denote the assignment weight between feature $f(x)_i$ and prototype $p_j$. This formulation allows multiple prototypes to contribute jointly to each image region, capturing finer-grained patterns that may appear in medical data.

ProtoPool reuses prototypes for related structures, such as similar tissue textures across different MRI slices. This pooling-based design helps maintain interpretability while providing a richer representation of complex anatomical variations. Since prototypes are no longer strictly tied to a single image patch, ProtoPool explanations may be less localized than those of PPNet or EPPNet, and it offers greater generalization. capacity.

\subsection{Faithfulness Evaluation Metrics}

To assess the interpretability and reliability of generated MRI images, we employ a set of quantitative metrics that evaluate both perceptual quality and faithfulness to the original training data. The metrices include Peak Signal-to-Noise Ratio (PSNR) \cite{PSNR}, Structural Similarity Index (SSIM) \cite{SSIM}, Learned Perceptual Image Patch Similarity (LPIPS) \cite{LPIPS}, and Faithfulness Score derived from prototype-based influence estimation. These metrics together can provide a comprehensive evaluation of image reconstruction quality.

\subsubsection{Peak Signal to Noise Ratio (PSNR)}

PSNR measure the fidelity between a generated image $\hat{x}$ and its ground-truth counterpart $x$. It quantifies the reconstruction error in terms of pixel-wise differences. The PSNR value is computed as:

\begin{equation}
\text{PSNR} = 10 \cdot \log_{10} \left( \frac{L^2}{\text{MSE}} \right),
\end{equation}
where $L$ denotes the maximum possible pixel intensity value (for normalized images, $L=1$), and the mean squared error (MSE) between two images is defined as:
\begin{equation}
\text{MSE} = \frac{1}{N} \sum_{i=1}^{N} (x_i - \hat{x}_i)^2,
\end{equation}
where $N$ is the total number of pixels. Higher PSNR indicates smaller reconstruction errors and higher image fidelity. Although PSNR provides a simple and interpretable measure, it is sensitive to minor pixel variations and may not fully reflect perceptual quality.

\subsubsection{Structural Similarity Index (SSIM)}

SSIM evaluates image quality based on structural consistency rather than absolute pixel differences. It measures the similarity between the luminance, contrast, and structure of two images, defined as:
\begin{equation}
\text{SSIM}(x, \hat{x}) = \frac{(2\mu_x \mu_{\hat{x}} + C_1)(2\sigma_{x\hat{x}} + C_2)}{(\mu_x^2 + \mu_{\hat{x}}^2 + C_1)(\sigma_x^2 + \sigma_{\hat{x}}^2 + C_2)},
\end{equation}
where $\mu_x$ and $\mu_{\hat{x}}$ are the mean intensities, $\sigma_x^2$ and $\sigma_{\hat{x}}^2$ are the variances, and $\sigma_{x\hat{x}}$ is the covariance between $x$ and $\hat{x}$. Constants $C_1$ and $C_2$ stabilize the division against weak denominators. SSIM values range from 0 to 1, with 1 indicating perfect structural similarity. This metric aligns more closely with human visual perception and is particularly suitable for medical imaging where structural integrity is critical.

\subsubsection{Learned Perceptual Image Patch Similarity (LPIPS)}

While PSNR and SSIM rely on pixel-wise or low-level features, the Learned Perceptual Image Patch Similarity (LPIPS) metric uses deep feature representations from pretrained convolutional networks to estimate perceptual differences. Let $\phi_l(x)$ and $\phi_l(\hat{x})$ denote the activation maps extracted from layer $l$ of a pretrained network (e.g., VGG or AlexNet). The LPIPS distance between $x$ and $\hat{x}$ is computed as:
\begin{equation}
\text{LPIPS}(x, \hat{x}) = \sum_{l} w_l \| \hat{\phi}_l(x) - \hat{\phi}_l(\hat{x}) \|_2^2,
\end{equation}
where $\hat{\phi}_l$ denotes normalized feature maps and $w_l$ are learned weights that capture the relative importance of each layer. Lower LPIPS values indicate higher perceptual similarity. LPIPS captures subtle structural and textural differences, often overlooked by traditional measures, making it relevant for assessing the realism of diffusion-generated MRI images.

\subsubsection{Faithfulness Score}

The Faithfulness Score quantifies the consistency between the influence estimated by prototypes and the actual contribution of training examples to a generated image. It measures how well the prototype-based explanation aligns with the underlying diffusion generation dynamics.

Given a set of normalized influence scores $\{ NIS_j(x) \}_{j=1}^{m}$ for a generated image $x$, where $NIS_j(x)$ represents the relative contribution of prototype $p_j$, the faithfulness score is defined as:
\begin{equation}
F(x) = \frac{1}{m} \sum_{j=1}^{m} NIS_j(x) \cdot \text{corr}(p_j, x),
\end{equation}

where $\text{corr}(p_j, x)$ denotes the spatial correlation between prototype $p_j$ and the corresponding region in $x$. The resulting correlation score lies in $[0,1]$, where higher values indicate more faithful and consistent explanations.

The metric provides an intrinsic measure of interpretability, and links the visual and statistical consistency of prototype activations with the diffusion process. In medical imaging, high faithfulness score implies the model explanations reliably represent anatomical patterns influenced the generation of synthetic MRI images, hence strengthens the trustworthiness of the generative framework.

\section{Results and Discussion}

\subsection{Dataset}

The experiments were performed on the DUKE Breast MRI dataset \cite{dce_mri2018}. It is a popular breast MRI images dataset includes different types of breast tissue, lesion variations, and anatomical differences, which makes it a good choice for evaluating generative models in medical imaging. 11{,}860 high-resolution MRI images were selected from the dataset to train a diffusion model for single class image generation. The model trained in a supervised way using anatomical segmentation maps as guidance to maintain proper spatial structure and more realistic generated images.

\begin{figure}[t]
\centering
\includegraphics[scale=0.30]{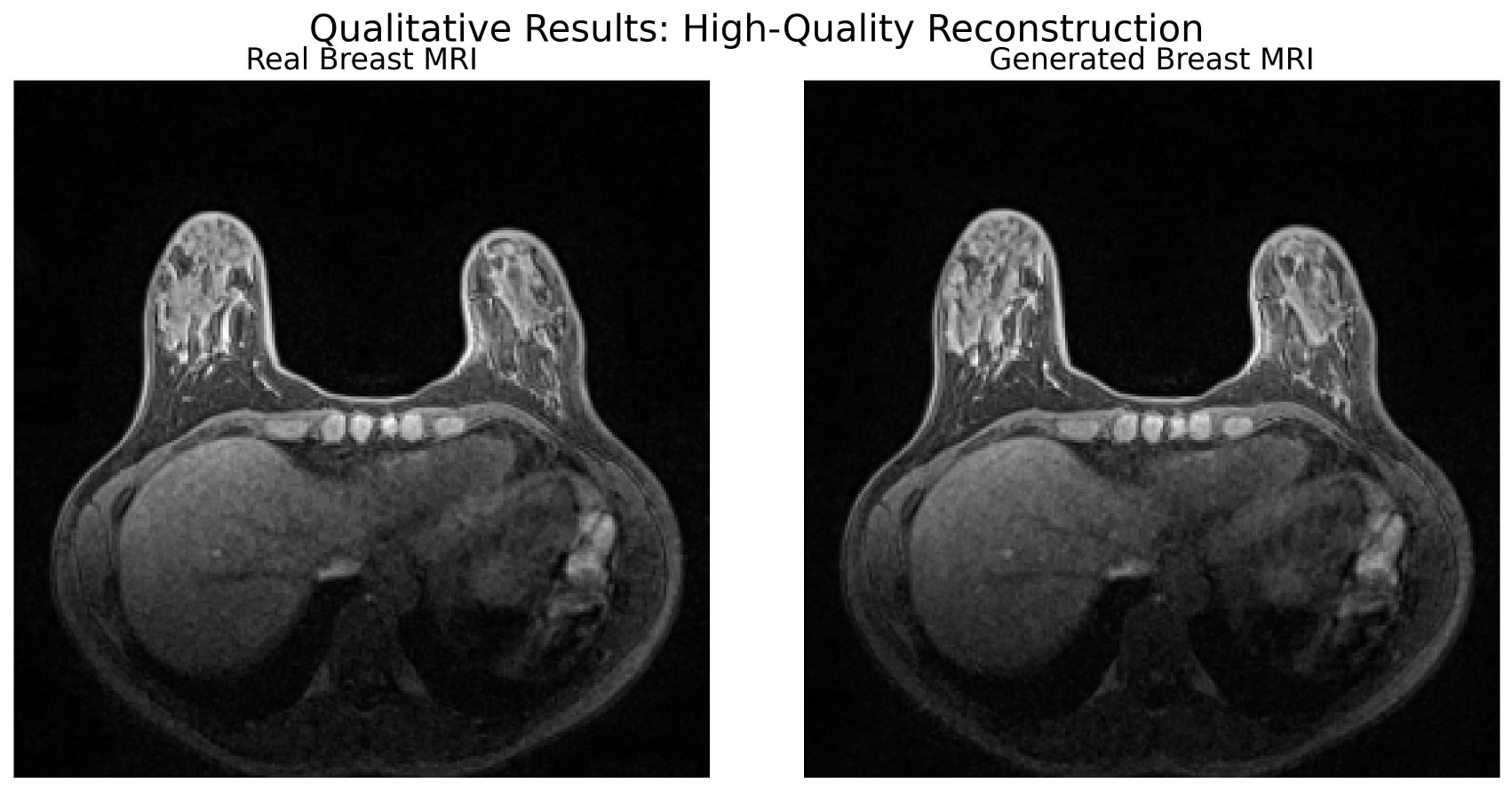}
\caption{Comparison of real and synthetic breast MRI images generated with Diffusion Model.}
\label{fig:real_vs_fake}
\end{figure}

The synthetic breast MRI images were evaluated by comparing them with real images. The observed  Frechet Inception Distance (FID) \cite{FID} score was 1.39, showing that the generated images are very close to real ones in terms of features. Another measure, the segmentation-based Dice similarity score was observed 0.9478, which means the anatomical structures were well preserved in the generated images. The results show that the diffusion model is able to capture fine details and structure of breast MRI images while maintaining high fidelity and anatomical consistency, as shown in Figure \ref{fig:real_vs_fake}.

\subsection{Experimental Setup}
Experiments were performed on NVIDIA GeForce RTX 2070 GPUs for training and evaluation, each of which has 8GB of memory. The system runs on the Ubuntu Linux operating system, providing a stable and efficient computing environment for deep learning experiments. All experiments are carried out in a virtual environment to ensure reproducibility and keep dependencies separate.

Model training is performed using batch processing to optimize memory utilization of both GPUs. The virtual environment is configured with specific package versions to ensure the compatibility and reproducibility of the results. All hyperparameters are carefully tuned based on preliminary validation experiments to achieve optimal performance while maintaining computational efficiency.

Specifically, our model configuration was tailored to the constraints of our low-data medical imaging task. We set the generated image resolution to 256x256 pixels, which maintains a balance between capturing sufficient anatomical detail for breast MRI analysis and maintaining computational feasibility. The model was trained for 200 epochs to ensure adequate learning without overfitting. Employed a relatively low learning rate of 2e-5, which has been empirically validated to provide stable training dynamics for diffusion models.

\subsection{Denoising Trajectory and Explainability Observation}

To analyze the internal behavior of the trained diffusion model, we performed a stepwise visualization of the denoising process using our own model. The objective is to understand how the model reconstructs anatomical structures from pure noise over multiple diffusion timesteps.

\begin{figure}[t]
\centering
\includegraphics[scale=0.30]{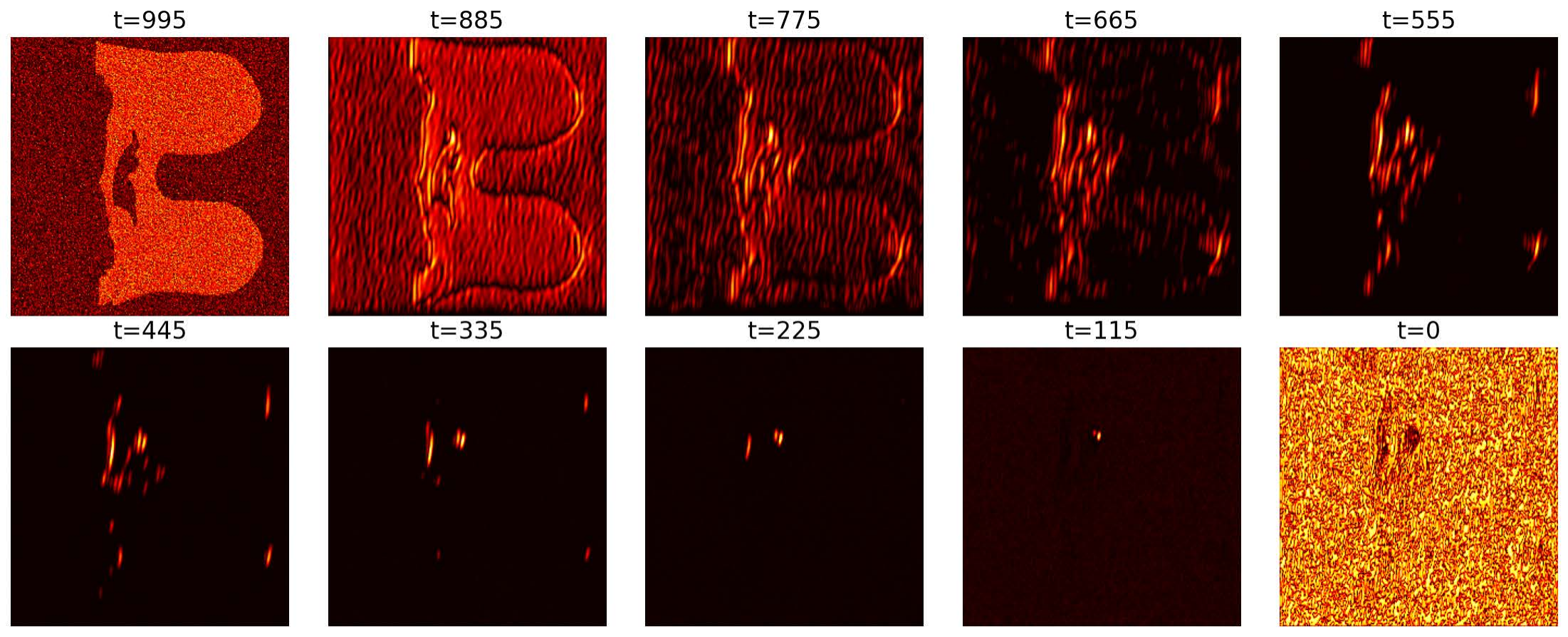}
\caption{Visualization of the denoising trajectory across diffusion timesteps.}
\label{fig:noise_prediction}
\end{figure}

The experiment began with a randomly initialized noise image concatenated with a binary segmentation mask representing the breast region. At the initial step, the input consisted of pure Gaussian noise with no discernible structure. As denoising progressed, the model iteratively refine the noise distribution, with anatomical outlines and intensity patterns gradually emerging. By the final timestep ($t = 995$), the generated image displayed a coherent breast MRI structure consistent with real images from the dataset. To interpret this process, we examined the noise prediction maps at each stage, representing the magnitude and spatial distribution of estimated noise. Early stages indicates the high-intensity noise, while later stages revealed localized suppression along tissue boundaries and glandular regions. This progressive reduction in noise magnitude illustrates how the model focuses on anatomically relevant areas, transforming unstructured noise into realistic medical imagery as shown in (Figure \ref{fig:noise_prediction}). Each frame in the figure represents the predicted noise magnitude at a specific stage of the generative process. The progression demonstrates how the trained diffusion model gradually reconstructs the anatomical features of breast MRI images through iterative denoising.

\subsection{Quantitative Analysis and Prototype Explainability}

\begin{figure}[t]
\centering
\includegraphics[scale=0.30]{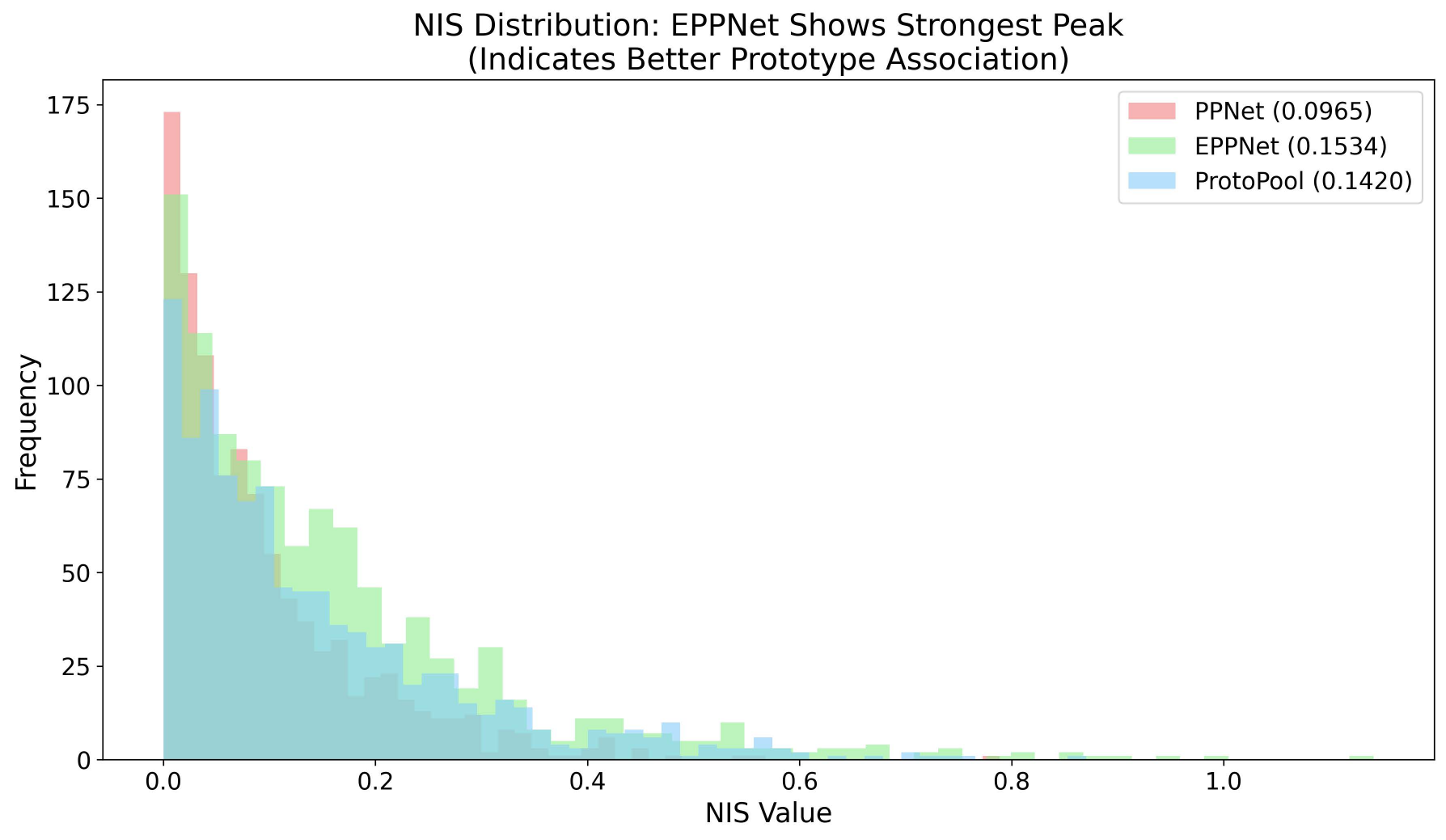}
\caption{Distribution of Normalized Influence Scores (NIS) across generated samples.}
\label{fig:nis}
\end{figure}

To evaluate the interpretability and quantitative performance of the trained diffusion model, three prototype based explainability architectures PPNet, EPPNet, and ProtoPool, were applied to the generated breast MRI images. These models were used to compute Normalized Influence Scores (NIS).  The figure \ref{fig:nis}, shows how prototype activations vary for PPNet, EPPNet, and ProtoPool. Plot shows how strongly each prototype from the training data contributes to the generation of a specific synthetic image. The faithfulness score derived from these influence distributions provides a quantitative measure of how accurately the prototype based explanations align with the actual generative process. Quantitative evaluation of our generated breast MRI images is provided in Table \ref{tab:image_quality}

This is clear that EPPNet maintains a more balanced and well-separated influence across its prototypes.

Across all models, the generated breast MRI images exhibited consistent visual quality, supported by high quantitative metrics. The diffusion model achieved a Peak Signal-to-Noise Ratio (PSNR) of $19.37 \pm 1.67$, a Structural Similarity Index (SSIM) of $0.6530 \pm 0.1052$, and a Learned Perceptual Image Patch Similarity (LPIPS) of $0.2893 \pm 0.1050$. These results confirm that the generated images maintain both high fidelity and perceptual realism relative to the real dataset.

\begin{table}[t]
\centering
\caption{Quantitative evaluation of generated breast MRI images using standard image quality metrics.}
\label{tab:image_quality}
\begin{tabular}{l c c c}
\hline
\textbf{Metric} & \textbf{Mean} & \textbf{SD} & \textbf{Observation} \\
\hline
PSNR  & 19.37 & $\pm$ 1.67 & Stable pixel fidelity \\
SSIM  & 0.6530 & $\pm$ 0.10 & Good structural match \\
LPIPS & 0.2893 & $\pm$ 0.10 &  Strong perceptual quality \\
\hline
\end{tabular}
\end{table}

\begin{figure}[t]
\centering
\includegraphics[scale=0.30]{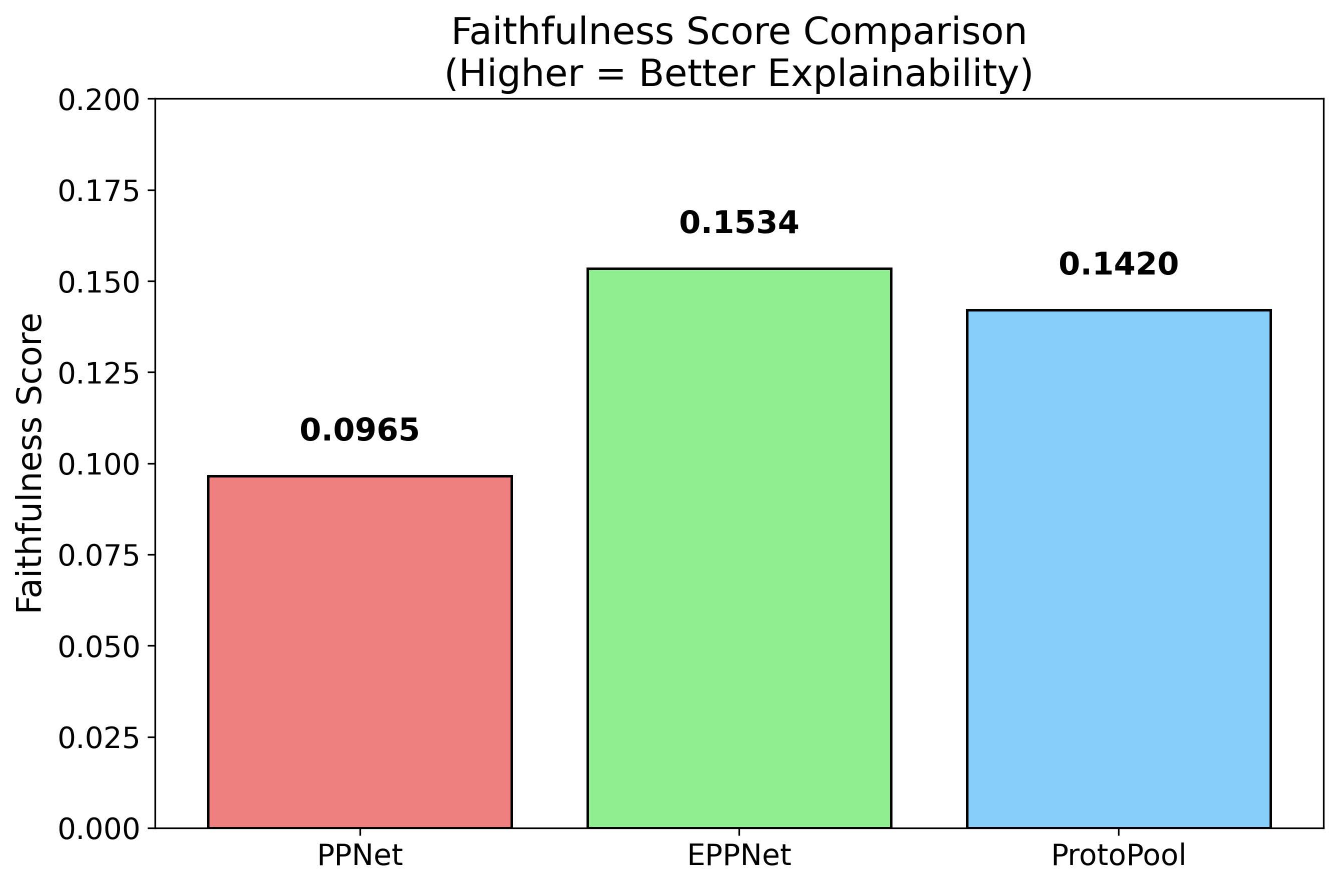}
\caption{Comparison of faithfulness scores for PPNet, EPPNet, and ProtoPool on generated breast MRI images.}
\label{fig:faith}
\end{figure}

The comparative faithfulness analysis in Figure \ref{fig:faith} shows that EPPNet achieved the highest faithfulness score of $0.1534$, followed by ProtoPool with $0.1420$, and PPNet with $0.0965$. The superior performance of EPPNet reflects its enhanced prototype normalization and diversity constraints, which allow it to provide more reliable and interpretable associations between real and synthetic image regions. ProtoPool shows a strong generalization by dynamically reusing prototypes across structurally similar features, while PPNet, although simpler, produced more localized explanations.

Overall, the results demonstrate that the proposed diffusion-based framework not only produces high-fidelity synthetic breast MRI images but also provides faithful and transparent explanations of the generative process. This combination of visual quality and explainability marks a promising step toward trustworthy generative modeling in medical imaging.

\subsection{Discussion}

The step-by-step denoising process clearly shows how useful anatomical details gradually come out from pure noise, giving a visual understanding of how the model works. This also suggests that explainability can be studied directly from the internal process of diffusion models, instead of depending only on post hoc explanation methods.

The quantitative analysis show that the proposed model keeps a good balance between image quality and interpretability. The PSNR, SSIM, and LPIPS scores indicate that the generated images are close to real MRI data both at pixel level and in visual appearance. Among the prototype-based methods, EPPNet gives the most consistent and reliable explanations, suggesting that normalization and diversity of prototypes help improve interpretability in single-class medical imaging tasks. ProtoPool provides similar performance showing that flexible sharing prototypes can handle small variations in anatomical structures.

\section{Conclusion}\label{sec:con}
This work is a step forward in building more reliable systems for medical imaging. This study shows that diffusion models can generate realistic and anatomically accurate MRI images. Combining with the diffusion process with prototype-based methods, we can get better understand how the diffusion model generates images and which parts of the training data influence them. The results show that EPPNet gives the most reliable explanations, showing that diffusion models can be both effective and transparent. Overall, the explainability and image quality are not opposite goals, but actually work well together in building trustworthy models. Hence, by studying how prototypes contribute to image generation and how the image improves step by step during denoising, we can better understand how the model makes decisions. This combination of transparency and realism gives a strong direction for using generative AI in medical research and diagnostic support systems.

\bibliographystyle{plain}
\fontsize{10}{10}\selectfont
%\addcontentsline{toc}{section}{\textbf{REFERENCES}}
\renewcommand{\bibname}{REFERENCES}
\bibliography{references}
%\end{spacing}

\end{document}